\documentclass[a4paper,fleqn]{cas-sc}

\usepackage[numbers,sort&compress]{natbib}

\def\tsc#1{\csdef{#1}{\textsc{\lowercase{#1}}\xspace}}
\tsc{WGM}
\tsc{QE}
\tsc{EP}
\tsc{PMS}
\tsc{BEC}
\tsc{DE}

\begin{document}
\let\WriteBookmarks\relax
\def\floatpagepagefraction{0.7}
\def\textpagefraction{.1}

\shorttitle{Spiking Transformers as a Zero-Temperature Limit}

\shortauthors{Sotero et al.}

\title[mode = title]{Softmax-Free Attention in Spiking Transformers as the\\ Zero-Temperature Limit of Two-Level Systems}            

\author[1,2,3]{Roberto C. Sotero}[type=editor,
    auid=000,bioid=1,
    orcid=0000-0001-9640-2722]
\cormark[1]
\ead{roberto.soterodiaz@ucalgary.ca}
\credit{Conceptualization, Funding, Methodology, Software, Writing}

\author[1,2, 3]{Pouya Taghipour}
\credit{Methodology, Writing}

\affiliation[1]{organization={Department of Radiology, University of Calgary},
    city={Calgary},
    state={AB},
    country={Canada}}

\affiliation[2]{organization={Department of Biomedical Engineering, University of Calgary},
    city={Calgary},
    state={AB},
    country={Canada}}

\affiliation[3]{organization={Hotchkiss Brain Institute, University of Calgary},
    city={Calgary},
    state={AB},
    country={Canada}}
    
\author[4]{Jose M. Sanchez-Bornot}
\credit{Methodology, Writing}
\affiliation[4]{organization={Intelligent Systems Research Center, Ulster University},
    city={Derry Londonderry},
    state={Northern Ireland},
    country={United Kingdom}}

\cortext[cor1]{Corresponding author}

\begin{abstract}
Spiking Transformers achieve energy efficiency by discarding softmax normalization in favor of thresholded, binary attention. However, this removal breaks the thermodynamic framework used to analyze attention networks, as the normalized measure and partition function cease to exist. We restore this theoretical foundation by proving that softmax-free spiking attention operates exactly as the zero-temperature limit of a product of independent two-level systems. This correspondence places spiking attention strictly in the frozen ordered phase of the thermodynamic phase diagram for all inputs. Consequently, the entropic cost of organizing attention collapses, replaced by a Landauer-type physical floor governed by action-potential generation. Furthermore, this ordered-phase pinning mathematically explains the optimization difficulty caused by vanishing surrogate gradients. We validate this theory with forward-relaxation experiments, placing both standard and neuromorphic attention mechanisms on a unified thermodynamic axis.
\end{abstract}

\begin{highlights}
\item Spiking attention equals the zero-temperature limit of two-level systems
\item Ordered-phase pinning makes spiking attention sparse yet hard to optimise
\item The entropic cost collapses to a Landauer floor set by spike generation
\item The surrogate slope acts as a backward temperature controlling trainability
\item Experiments place spiking and softmax attention on one thermodynamic axis
\end{highlights}


\maketitle

\section{Introduction}
\label{sec:intro}
Spiking neural networks have gained prominence in machine learning as the energy-efficient, event-driven counterpart to standard deep networks \cite{yao2023spikedriven, zhou2022spikformer, shi2024towards}. To adapt the Transformer \cite{vaswani2017attention} to this framework, recent models encode queries, keys, and values using spiking neurons, replacing continuous floating-point attention mechanisms with spike-based computation \cite{zhou2022spikformer, zhu2023spikegpt, zhou2026winnertakeall}. However, the standard softmax operation is fundamentally ill-suited to this scheme since its exponentials and divisions produce dense, real-valued weights that conflict with the addition-only processing required for high energy efficiency \cite{yao2023spikedriven}. Softmax is also unnecessary here, as binary queries and keys naturally yield non-negative, sparse scores \cite{zhou2022spikformer}. Consequently, the removal of softmax has become a defining feature of these architectures, often reducing the resulting operator to a form of linear attention \cite{yao2023spikedriven, zhou2026winnertakeall}.

Concurrently, statistical-mechanical models of softmax attention have been developed \cite{ramsauer2021, krotov2016}. The attention weights of scaled dot-product attention form a Boltzmann measure over keys, with the scaled scores acting as negative energies \cite{ramsauer2021}. A recently proposed thermodynamic framework builds on this identity to analyse attention training, deriving a conservation-law decomposition of the loss-reduction budget and locating a characteristic temperature that separates ordered from disordered attention \cite{soterothermo2026}. However, this framework relies entirely on the softmax denominator, which acts as the partition function and provides the normalized measure. By discarding the softmax operation, spiking Transformers eliminate both, leaving the characteristic temperature without a direct analogue. Consequently, the existing framework cannot be applied to spiking architectures, despite the fact that their primary motivation, minimizing energy consumption, is precisely the type of problem a thermodynamic theory should resolve.

In this paper, we address this issue by showing that softmax-free spiking attention, in which each query--key score is thresholded into a spike, is the zero-temperature limit of a product of independent two-level systems \cite{kittel1980thermal}. This recovers a statistical-mechanical description precisely where the softmax--Boltzmann identity fails. We state explicitly which findings are exact mathematical consequences of the construction and which are testable predictions. Our contributions are as follows:
\begin{itemize}
\item We establish an explicit mapping between a product of independent two-level systems and softmax attention. Under this mapping, each energy gap is determined by the scaled score relative to the firing threshold, and the logistic surrogate gradient becomes the derivative of the corresponding occupation probability at a finite backward temperature $1/\alpha$.
\item Spiking attention is pinned in the forward pass to a fully ordered, zero-entropy state for every input, while gradients predominantly update pairs whose scores lie within a window of $\sim 1/\alpha$ around the threshold. The spike nonlinearity thus guarantees sparse, binary activity while confining score updates to this narrow band.
\item The per-input entropic term that bounds the loss reduction of softmax attention \cite{soterothermo2026} vanishes, and we postulate in its place a Landauer-type floor \cite{landauer1961} set by the change in entropy of the across-input spike pattern.
\item We experimentally validate the proposed theoretical mapping on CIFAR-10 \cite{krizhevsky2009cifar}, placing both a spiking and a softmax-based Transformer on a common thermodynamic axis to confirm the predicted entropy dynamics and temperature-dependent trainability. 
\end{itemize}

\section{Related Work}
\label{sec:related}

\subsection{Spiking Transformers.}
The adaptation of the Transformer \cite{vaswani2017attention} to spiking neural networks relies fundamentally on removing the softmax function. Softmax requires floating-point exponentiation and division, operations that violate the multiplication-free, binary-spike paradigm of neuromorphic hardware \cite{roy2019towards, davies2018loihi}. To address this, Spikformer introduced a spiking self-attention mechanism that operates exclusively on spike-form queries, keys, and values without a softmax layer \cite{zhou2023spikformer}. Similarly, the Spike-driven Transformer collapsed the query-key-value interaction into masking and sparse addition. This produced a multiplication-free, effectively linear attention with an order-of-magnitude reduction in computational energy \cite{yao2023spikedriven}, which was later generalized into a meta-architecture for neuromorphic deployment \cite{yao2024spikedrivenv2}. Subsequent works have extended this operator to hierarchical attention \cite{zhou2024qkformer} and language models \cite{zhu2024spikegpt}. While energy efficiency motivates these architectures, the literature typically quantifies these gains simply by counting hardware-level operations, specifically, noting the energy difference between multiply-accumulate (MAC) operations and simple accumulations \cite{horowitz20141}.

\subsection{Energy-based and thermodynamic views of attention.}
The connection between attention mechanisms and statistical physics is well established. Modern Hopfield networks frame self-attention as an energy-based associative memory where the retrieval step maps exactly to the softmax operation, and the softmax denominator serves as the partition function \cite{ramsauer2021,krotov2016}. Concretely, for a query $\mathbf{q}_i\in\mathbb{R}^{d_k}$ and keys $\{\mathbf{k}_j\}_{j=1}^N$, with scores and energies $s_{ij}=\mathbf{q}_i^\top \mathbf{k}_j/\sqrt{d_k}$ and $E_{ij}=-s_{ij}$, scaled dot-product attention at inverse temperature $\beta=1/T$ is the Boltzmann measure:
\begin{equation}
\begin{split}
A_{ij} &= \frac{e^{-\beta E_{ij}}}{Z_i} \\
Z_i &= \sum_{l=1}^N e^{-\beta E_{il}}
\end{split}
\label{eq:softmax}
\end{equation}
recovering standard attention at $\beta=1$. Under this identity the row vector $\mathbf{A}_i$ is a probability vector with Shannon entropy $S_i=-\sum_j A_{ij}\ln A_{ij}\in[0,\ln N]$, order parameter $\phi=1-\langle S_i\rangle/\ln N$ (where $\langle \cdot \rangle$ denotes the average over all queries), and free energy $F_i=-T\ln Z_i$, all exact functionals of Equation~\eqref{eq:softmax}. A recent thermodynamic framework extended this identity to model attention training \cite{soterothermo2026}. This approach decomposed the loss-reduction budget into heat and work, derived a Landauer-type bound $-\Delta\mathcal{L}\ge T\vert{}\Delta\langle S\rangle\vert{}$, and identified a characteristic temperature $T_c$ that separates the disordered and ordered phases. The framework relies entirely on the partition function $Z_i$ and the normalized measure in Equation~\eqref{eq:softmax}, leaving the softmax-free architectures discussed above unaddressed. By characterizing the statistical-mechanical structure of spiking attention, the present work fills this theoretical gap.

\subsection{Two-level systems and spin models of neural computation.}
Mapping neural state variables onto interacting spins is a long-standing programme in statistical physics. Little \cite{little1974} modelled neuronal firing as a two-state Ising variable and identified persistent network states with the ordered phase of the resulting spin system. Building on this, Hopfield cast associative memory as relaxation on an energy landscape with symmetric pairwise couplings \cite{hopfield1982neural}, and the storage capacity and phase structure of such networks were later obtained using spin-glass methods \cite{amit1985spin}. The elementary unit shared by these models is the two-level system. 

Consider a two-level system \cite{kittel1980thermal,pathria2011statistical} with energy states $0$ and $\Delta$. At an inverse temperature $\beta=1/T$, its upper-state occupation probability $p$ and partition function $z$ are given by:
\begin{equation}
\begin{split}
p(\beta) &= \frac{e^{-\beta\Delta}}{1+e^{-\beta\Delta}} = \frac{1}{1+e^{\beta\Delta}} \\
z(\beta) &= 1+e^{-\beta\Delta}
\end{split}
\label{eq:tls_occ}
\end{equation}
where $p(\beta)$ corresponds to the Fermi-Dirac distribution, which is mathematically identical to the logistic sigmoid $\sigma(x)=(1+e^{-x})^{-1}$ via the relation $p(\beta)=\sigma(-\beta\Delta)$.

Given the mean energy $\langle E\rangle=\Delta\,p(\beta)$ and the binary entropy $\mathcal{H}_{bin}(p)=-p\ln p-(1-p)\ln(1-p)$, the heat capacity $c$ takes the well-known Schottky form \cite{kittel1980thermal}:
\begin{equation}
c(\beta)=\frac{\partial\langle E\rangle}{\partial T}=\beta^{2}\Delta^{2}\,p(\beta)\big(1-p(\beta)\big)
\label{eq:tls_cv}
\end{equation}

Two asymptotic limits govern the behaviour of this system. As $T\to0$, the system freezes into its minimal-energy state. In this limit, $p$ approaches either $0$ or $1$ depending on the sign of the gap, the entropy vanishes, and the order parameter $1-\mathcal{H}_{bin}(p)/\ln2$ tends to $1$. As $T\to\infty$, the states become equally occupied ($p\to\tfrac12$) and the order parameter tends to $0$. The heat capacity in Equation~\eqref{eq:tls_cv} vanishes in both limits and peaks at an intermediate temperature near $\beta\Delta\approx2.4$. This peak constitutes the Schottky anomaly, the defining thermodynamic fingerprint of a two-level system \cite{kittel1980thermal}. Crucially, this elementary unit is already native to energy-based neural models. In the Boltzmann machine, a binary neuron takes a value of one with a probability equal to the logistic function of its input \cite{ackley1985learning}. Following Equation~\eqref{eq:tls_occ}, this is exactly a two-level occupation, where taking a deterministic readout physically places that unit at the zero-temperature limit.

\section{Methods}
\label{sec:method}

\subsection{Spiking attention baseline.}
A spiking neuron with threshold $\vartheta$ is defined by the Heaviside step function $\sigma_\vartheta(x)=\Theta(x-\vartheta)$. Because its true derivative vanishes almost everywhere, the backward pass relies on a surrogate derivative $\sigma_\alpha'$, where $\sigma_\alpha(x)=(1+e^{-\alpha x})^{-1}$ is the logistic sigmoid with slope $\alpha>0$ \cite{neftci2019surrogate}. Given a pre-spike scaling factor $\gamma>0$ and the scores $s_{ij}$ from Equation~\eqref{eq:softmax}, the spiking attention matrix and its corresponding output are formulated as:
\begin{align}
a_{ij} &= \Theta(\gamma s_{ij}-\vartheta) \in \{0,1\} \label{eq:spk_a} \\
\mathbf{o}_i &= \sum_{j=1}^N a_{ij} \, \mathbf{v}_j \label{eq:spk_o}
\end{align}
Equations~\eqref{eq:spk_a} and \eqref{eq:spk_o} capture the core mechanism of the dominant architectures in the spiking transformer literature \cite{zhou2023spikformer,yao2023spikedriven,yao2024spikedrivenv2,zhou2024qkformer,zhu2024spikegpt}. The critical feature of this formulation is that an independent, per-element threshold replaces the cross-key normalization of softmax, thereby freeing the row vector $\mathbf{a}_i$ from any stochastic constraint.

\subsection{The breakdown of the softmax thermodynamic framework.}
Because the vector $\mathbf{a}_i$ carries no row-stochastic constraint, removing the softmax operation is not merely a change of temperature within Equation~\eqref{eq:softmax}. For generic scores, the activations $\{a_{ij}\}_j$ are not the Boltzmann measure of any finite inverse temperature. A finite-$\beta$ measure $e^{-\beta E_{ij}}/Z_i$ is strictly positive and sums to one, whereas $a_{ij}\in\{0,1\}$ vanishes exactly for every key with $\gamma s_{ij}<\vartheta$, and the row sums to $\sum_j a_{ij}\in\{0,\dots,N\}$ rather than to one. No finite $\beta$ and normalizer $Z_i$ can reproduce a vector with exact zeros or with a mass different from one. The partition function, the free energy $-T\ln Z_i$, and the fluctuation identities $\chi=\operatorname{Var}(S_i)$ and $C_v=\operatorname{Var}(E)/T^2$, defined only relative to such a measure, are therefore undefined for spiking attention. Row-normalizing the pattern does not repair this: it returns the uniform distribution on the supra-threshold support. This constitutes an infinite-temperature measure on a support that is itself selected at zero temperature, which lies entirely outside the one-parameter family of Equation~\eqref{eq:softmax}.

\subsection{The two-level Boltzmann correspondence.}
To model these non-stochastic activations, we propose to change the ensemble rather than the temperature. Softmax attention is the partition function of a single coupled $N$-state categorical system, with the keys competing through the shared normalizer $Z_i$ as in a Potts model \cite{krotov2016}. Deleting the softmax layer removes that shared denominator and, with it, the coupling between keys, leaving one independent two-level decision per query-key pair. We assign to each pair $(i,j)$ a two-level system with energies $\{0,\Delta_{ij}\}$ and gap $\Delta_{ij}=\vartheta-\gamma s_{ij}$. Its spiking-state occupation and partition function at inverse temperature $\beta_s$ are:
\begin{align}
p_{ij}(\beta_s) &= \frac{e^{-\beta_s\Delta_{ij}}}{1+e^{-\beta_s\Delta_{ij}}}
   = \sigma\!\big(\beta_s(\gamma s_{ij}-\vartheta)\big) \label{eq:twolevel_p} \\
z_{ij}(\beta_s) &= 1+e^{-\beta_s\Delta_{ij}} \label{eq:twolevel_z}
\end{align}
where the occupation is the logistic form of the Fermi-Dirac distribution from Equation~\eqref{eq:tls_occ}. By defining the net input to the threshold as $u_{ij} = \gamma s_{ij} - \vartheta = -\Delta_{ij}$, we observe that $\sigma(\beta_s u_{ij}) \to \Theta(u_{ij})$ pointwise as $\beta_s \to \infty$ for $u_{ij} \neq 0$. Consequently, the occupation freezes onto the hard threshold:
\begin{equation}
\lim_{\beta_s\to\infty} p_{ij}(\beta_s)=\Theta(\gamma s_{ij}-\vartheta)=a_{ij}
\label{eq:correspondence}
\end{equation}
Therefore, the spike-form operator of Equation~\eqref{eq:spk_a} is exactly the zero-temperature ($T_s=1/\beta_s\to0$) limit of the product ensemble defined in Equation~\eqref{eq:twolevel_p}. The training surrogate is the same object evaluated at a finite temperature. Differentiating Equation~\eqref{eq:twolevel_p} and setting $\beta_s=\alpha$ yields the score-gradient:
\begin{equation}
\left.\frac{\partial p_{ij}}{\partial s_{ij}}\right|_{\beta_s=\alpha}
   =\alpha\gamma\,\sigma'\big(\alpha(\gamma s_{ij}-\vartheta)\big)
\label{eq:surrogate_grad}
\end{equation}
which is exactly the fast-sigmoid surrogate of slope $\alpha$. Thus, the forward pass is evaluated at $\beta_s=\infty$ and the gradients at the backward inverse temperature $\beta_s=\alpha$.

\begin{figure}[t!]
\centering
\includegraphics[width=\textwidth]{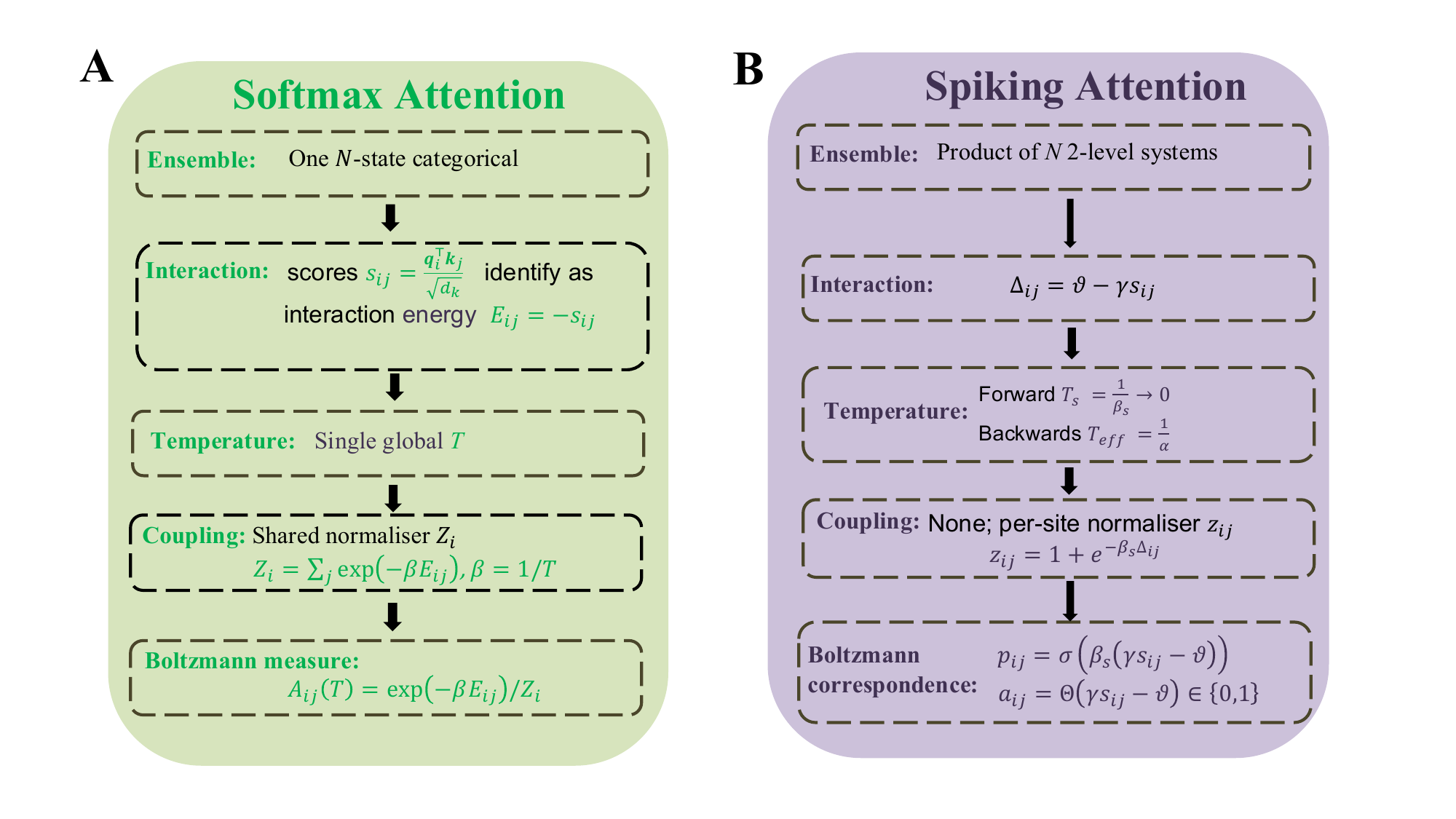}
\caption{The spiking-Boltzmann correspondence. Each column is read from top to bottom as a derivation, from the underlying ensemble to its Boltzmann measure. \textbf{A}: softmax attention. The $N$ keys attended to by a query form a single coupled $N$-state categorical system. The scores $s_{ij}=\mathbf{q}_i^{\top}\mathbf{k}_j/\sqrt{d_k}$ define the interaction energy $E_{ij}=-s_{ij}$, a single global temperature $T$ fixes the inverse temperature $\beta=1/T$, the shared normaliser $Z_i=\sum_j e^{-\beta E_{ij}}$ couples the keys, and the attention weights are the Boltzmann measure $A_{ij}(T)=e^{-\beta E_{ij}}/Z_i$ of Equation~\eqref{eq:softmax}, with operating point $\beta=1$. \textbf{B}: softmax-free spiking attention. Deleting the shared normaliser decouples the keys into a product of $N$ independent two-level systems, one per query-key pair, each with energy gap $\Delta_{ij}=\vartheta-\gamma s_{ij}$ and partition function $z_{ij}=1+e^{-\beta_s\Delta_{ij}}$. The single temperature splits into a forward value $T_s=1/\beta_s$ and a backward value $T_{eff}=1/\alpha$ set by the surrogate slope. The two-level occupation $p_{ij}=e^{-\beta_s\Delta_{ij}}/z_{ij}=\sigma(\beta_s(\gamma s_{ij}-\vartheta))$ is the counterpart of $A_{ij}$, and the deployed hard spike $a_{ij}=\Theta(\gamma s_{ij}-\vartheta)\in\{0,1\}$ is its zero-temperature limit $T_s\to0$ (Equation~\eqref{eq:correspondence}). Softmax attention therefore operates at a finite, entropy-bearing temperature, whereas spiking attention is the zero-temperature limit of the corresponding two-level Boltzmann measure.}
\label{fig:correspondence}
\end{figure}

Equation~\eqref{eq:correspondence} serves as the spiking counterpart to the softmax-Boltzmann identity of Equation~\eqref{eq:softmax}. Removing softmax does not merely delete a normalizer but replaces one coupled $N$-state categorical system with $N$ decoupled two-level systems, and splits the single softmax temperature into a forward value ($\beta_s\to\infty$) and a backward value ($\beta_s=\alpha$). Figure~\ref{fig:correspondence} sets out the resulting dictionary. Letting $\mathcal{H}_{bin}(p)=-p\ln p-(1-p)\ln(1-p)$ denote the binary entropy, the categorical entropy of Equation~\eqref{eq:softmax} is likewise replaced by a sum of binary entropies, and the free energy by a sum of two-level free energies. This identification is exact under Equation~\eqref{eq:spk_a} and the logistic surrogate.

\subsection{Consequence 1: Ordered-phase pinning, sparsity, and optimization.}
Evaluating the forward pass at the zero-temperature limit ($\beta_s \to \infty$) fundamentally dictates the macroscopic state of the attention pattern. As the system cools, the occupation probabilities $p_{ij}$ freeze almost entirely into the discrete states $\{0,1\}$. To quantify this phase, we define the single-site binary entropy $s^{(2)}_{ij}=\mathcal{H}_{bin}(p_{ij})$, its global mean $\bar S^{(2)}=\langle s^{(2)}_{ij}\rangle$, and the corresponding order parameter $\phi^{(2)}=1-\bar S^{(2)}/\ln2$. Because the binary entropy vanishes completely when the system becomes deterministic, this zero-temperature limit strictly pins the network to a fully ordered phase:
\begin{equation}
\begin{split}
\bar S^{(2)} &\to 0 \\
\phi^{(2)} &\to 1
\end{split}
\qquad (\beta_s\to\infty)
\label{eq:pinning}
\end{equation}
This phase pinning has two immediate physical manifestations. First, the frozen $\{0,1\}$ states produce strictly sparse, binary activations. This is the precise multiplication-free regime that grants spiking attention its extreme energy efficiency \cite{yao2023spikedriven}. Second, this structural rigidity severely constrains the learning dynamics. Because the true gradient of a frozen state is zero, optimization must rely entirely on the surrogate derivative, whose magnitude $\frac{d}{du}\sigma(\alpha u) = \alpha\,\sigma(\alpha u)\big(1-\sigma(\alpha u)\big)$ decays rapidly outside a narrow thermal band of width $O(1/\alpha)$ around the threshold. By operating so deeply within this rigid, vanishing-gradient regime, the system exhibits the exact conditions that thermodynamic frameworks of attention associate with severe optimization difficulty \cite{soterothermo2026}.

\subsection{Consequence 2: The thermodynamic cost of spike generation.}
For standard softmax attention, the First Law of Learning frames the loss reduction $-\Delta\mathcal{L}$ as the free-energy budget that drives the system, decomposed into heat and work \cite{soterothermo2026}:
\begin{equation}
-\Delta\mathcal{L} \approx \underbrace{T\big(\langle S \rangle_{init} - \langle S \rangle_{final}\big)}_{\text{heat}} + \underbrace{\Delta I(\hat Y;Y)}_{\text{work}}
\label{eq:first_law_softmax}
\end{equation}
where $T$ is the softmax temperature and the heat term represents the entropic cost of sharpening the attention distribution. This imposes a fundamental Landauer limit on learning, requiring $-\Delta\mathcal{L}\ge T|\Delta\langle S\rangle|$ \cite{soterothermo2026,landauer1961}. However, because the two-level order parameter $\phi^{(2)}$ of spiking attention is strictly pinned near one (Equation~\eqref{eq:pinning}), its per-input entropy change is practically zero ($|\Delta\langle S^{(2)}\rangle|\approx0$). Consequently, the heat term in Equation~\eqref{eq:first_law_softmax} collapses, reducing the standard Landauer bound to the trivial inequality $-\Delta\mathcal{L}\ge0$. The fundamental thermodynamic cost must therefore shift from sharpening a local probability distribution to structuring a global spike pattern.

We formalize this shift through a physical postulate: because the local per-input pattern is structurally frozen, the loss reduction $-\Delta\mathcal{L}$ must act instead as the free-energy budget for reorganizing the macroscopic spike distribution across the entire dataset. We define the across-ensemble entropy of this global pattern as:
\begin{equation}
S_{spk}=\frac{1}{|\mathcal{N}|}\sum_{n\in\mathcal{N}}\mathcal{H}_{bin}(\pi_n)
\label{eq:sspk}
\end{equation}
where $\pi_n$ is the firing probability of attention neuron $n$ evaluated over the dataset, and $\mathcal{N}$ is the set of all attention neurons across all heads and layers. By treating this global spike distribution as the information-bearing subsystem, we substitute $S_{spk}$ for $\bar{S}$ and assign the surrogate backward temperature $T_{eff}=1/\alpha$. The spiking First Law and its associated Landauer bound then become:
\begin{align}
-\Delta\mathcal{L} &\approx
   \underbrace{T_{eff}\big(S_{spk}^{\,init}-S_{spk}^{\,fin}\big)}_{\text{heat}}
   +\underbrace{\Delta I(\hat Y;Y)}_{\text{work}} \label{eq:first_law} \\
-\Delta\mathcal{L} &\ge T_{eff}\,|\Delta S_{spk}| \label{eq:landauer_bound}
\end{align}

The work term carries over unchanged from the softmax case. Crucially, this framework maps directly to physical energy consumption. Independent of synaptic weight magnitudes, the attention-layer energy obeys the activity floor $E_{attn}\ge C_{AP}\,\bar n$, where $\bar n$ is the mean attention-neuron spike count and $C_{AP}$ is the energetic cost of a single action potential. Note that $S_{spk}$ is physically distinct from $\bar S^{(2)}$. While the local per-input pattern is frozen in the ordered phase, the global across-ensemble pattern remains highly informative, and it is precisely this global entropy that network specialization reduces. Because Equations~\eqref{eq:first_law} and \eqref{eq:landauer_bound} rest entirely on this theoretical postulate, they are strictly falsifiable. We test this empirically in Section~\ref{sec:exp}.

\begin{table}[t!]
\centering
\caption{Experimental configuration. The surrogate slope $\alpha$ is fixed for the relaxation experiment (Figure~\ref{fig:relaxation}) and swept for the backward-temperature experiment (Figure~\ref{fig:surrogate}).}
\label{tab:params}
\begin{tabular}{lll}
\toprule
Quantity & Symbol & Value \\
\midrule
Dataset             &                       & CIFAR-10 \\
Image size          &                       & $32\times32\times3$ \\
Patch size          &                       & $4$ \\
Embedding width     & $d_{\text{model}}$    & $64$ \\
Attention heads     &                       & $4$ \\
Transformer blocks  &                       & $2$ \\
MLP expansion       &                       & $4$ \\
Timesteps           & $N_t$                 & $4$ \\
\midrule
Firing threshold    & $\vartheta$           & $1$ \\
Score gain          & $\gamma$              & $1$ \\
LIF threshold       & $v_{\text{th}}$       & $1$ \\
Membrane decay      &                       & $0.5$ \\
Surrogate slope, relaxation & $\alpha$      & $25$ \\
Surrogate slope, sweep      & $\alpha$      & $2$ to $100$ ($7$ levels) \\
\midrule
Optimiser           &                       & AdamW \\
Learning rate       &                       & $10^{-3}$ \\
Weight decay        &                       & $10^{-4}$ \\
Epochs              &                       & $100$ \\
Batch size          &                       & $128$ \\
Seeds, relaxation / sweep &                 & $5$ / $3$ \\
\midrule
Two-level inverse temperature & $\beta_s$             & $10^{-1}$ to $10^{2}$ ($18$ levels) \\
Softmax temperature     & $T$                   & $10^{-2}$ to $10^{1}$ ($18$ levels) \\
\bottomrule
\end{tabular}
\end{table}

\section{Numerical Experiments}
\label{sec:exp}
We validate the proposed thermodynamic correspondence on the CIFAR-10 dataset \cite{krizhevsky2009cifar} using a compact spiking Transformer. To maintain a tractable thermodynamic state space, the architecture is intentionally scaled down (see Table~\ref{tab:params} for the complete experimental configuration). Images are provided to the network via direct coding over $N_t$ time steps. The model employs the threshold attention mechanism of Equation~\eqref{eq:spk_a} alongside leaky integrate-and-fire neurons featuring a soft reset mechanism, and is trained using the surrogate gradient defined in Equation~\eqref{eq:surrogate_grad}. 

This model is benchmarked against a standard softmax Vision Transformer \cite{dosovitskiy2021vit} of identical width and depth, which employs GELU activations in its multi-layer perceptrons in place of the spiking neurons. Because the two architectures share identical patch projections, embedding dimensions, head counts, layer depths, and optimisers, any divergence in their thermodynamic behaviour is strictly attributable to the replacement of the softmax operation with a per-element threshold. The two-level order parameter $\phi^{(2)}$ and specific heat $C_v^{(2)}$ are computed from the net inputs $u_{ij}=\gamma s_{ij}-\vartheta$ on the held-out test set. Conversely, the softmax baseline is characterised by the categorical-entropy order parameter $\phi=1-\langle S_i\rangle/\ln N$ and the specific heat $C_v=\operatorname{Var}(E)/T^2$ from Equation~\eqref{eq:softmax}, evaluated across a continuous temperature sweep.

\subsection{Forward relaxation and ordered-phase pinning.}
\begin{figure}[t!]
\centering
\includegraphics[width=\textwidth]{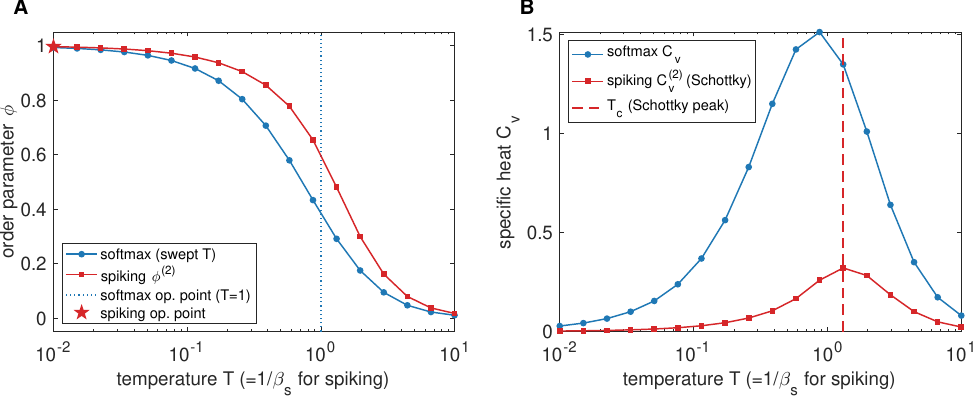}
\caption{Forward relaxation of the matched models on CIFAR-10. The trained spiking operator (red) is shown against the two-level inverse temperature $\beta_s$ and the softmax baseline (blue) against a swept temperature $T$, on a shared axis. \textbf{A}: order parameter. The softmax model sits at the disordering crossover at its native point $T=1$ (dotted line), while the spiking model is pinned near $\phi^{(2)}=1$ at its operating point $\beta_s\to\infty$ (star), validating the ordered-phase pinning of Equation~\eqref{eq:pinning}; the comparison concerns operating-point placement, not curve identity. \textbf{B}: specific heat. The spiking heat capacity shows an interior Schottky anomaly peaking at $T_c$ (dashed line), and the operating point at $T\to0$ lies well cold of $T_c$. The softmax peak is larger because it sums the energy variance over all $N$ keys, whereas $C_v^{(2)}$ is the heat capacity of one binary unit per pair.}
\label{fig:relaxation}
\end{figure}

Figure~\ref{fig:relaxation}A illustrates the order parameter as a function of temperature. For the softmax baseline, varying $T$ reveals a complete disordering crossover from an ordered regime at low temperatures to a uniform, maximum-entropy regime at high temperatures, with its native operating point ($T=1$) situated deep within the interior of this crossover. Conversely, the trained spiking model, evaluated along the two-level inverse temperature $\beta_s$, behaves fundamentally differently. At its native operating point ($\beta_s\to\infty$, indicated by a star), the order parameter is pinned near $\phi^{(2)}=1$, confirming the ordered-phase pinning predicted by Equation~\eqref{eq:pinning}. The trained attention operator maps every input into this fully ordered regime, wherein each query-key pair collapses onto a definite spiking decision and the local per-input entropy vanishes. Because the horizontal axis denotes the physical temperature $T$ for the softmax model and the two-level equivalent temperature $1/\beta_s$ for the spiking model, this comparison highlights the distinct thermodynamic operating regimes rather than a strict overlap of the curves. The softmax model operates at a finite, entropy-bearing temperature, whereas the spiking model operates strictly at zero temperature.

Figure~\ref{fig:relaxation}B presents the specific heat, confirming that the spiking attention mechanism acts as a decoupled two-level system. The specific heat of the spiking model rises from zero at low temperatures, passes through a single interior maximum at a critical temperature $T_c$, and decays back toward zero. This behaviour constitutes a Schottky anomaly, which is the classical thermodynamic signature of a two-state degree of freedom. A single coupled categorical system cannot produce this per-pair heat capacity profile. Its emergence serves as direct evidence that removing the softmax operator decouples the keys into independent two-level systems, exactly as Equation~\eqref{eq:correspondence} asserts. The spiking operating point ($1/\beta_s \to 0$) lies far on the cold side of $T_c$, deep within the frozen phase, which is entirely consistent with the order-parameter pinning observed in panel A. The macroscopic peak of the softmax model is naturally larger because it aggregates the energy variance across all $N$ competing keys, whereas $C_v^{(2)}$ represents the heat capacity of a single binary degree of freedom per pair. Thus, the two ensembles carry different intrinsic fluctuations, and their crossover temperatures need not coincide. The essential validation of the correspondence lies in the qualitative shape of the curve: an interior Schottky peak with the operating point anchored on its cold flank, irrespective of the relative heights of the two peaks.

\subsection{Trainability and the backward temperature.}
\begin{figure}[t!]
\centering
\includegraphics[width=\textwidth]{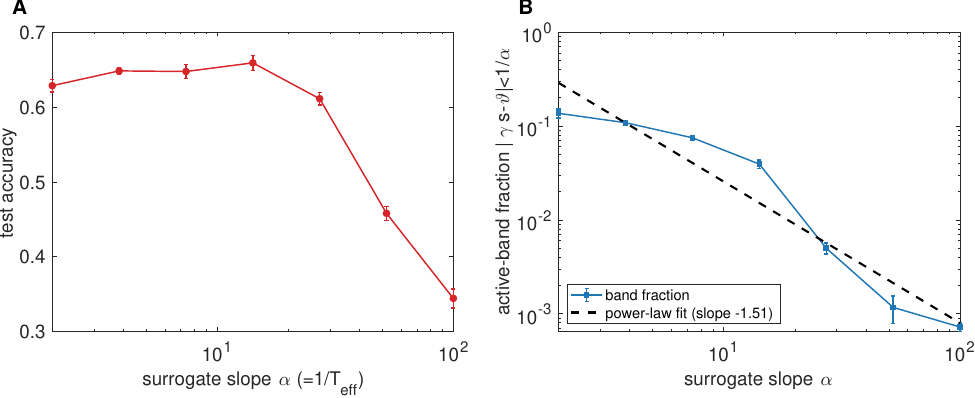}
\caption{Surrogate slope as the backward temperature on CIFAR-10. \textbf{A}: test accuracy against $\alpha=1/T_{\text{eff}}$, with error bars over seeds. Accuracy peaks across a broad intermediate plateau and collapses as $\alpha$ grows: large $\alpha$ is the cold backward temperature that starves the gradient, and the smallest $\alpha$ is the biased high-temperature side. \textbf{B}: active-band fraction, the fraction of pairs with $|\gamma s_{ij}-\vartheta|<1/\alpha$, against $\alpha$ on logarithmic axes; the power-law fit (slope in the legend) confirms the $O(1/\alpha)$ surrogate support of Equation~\eqref{eq:surrogate_grad}. The fraction is averaged over the attention array, so its across-seed spread is below the marker size. Panels A and B give mechanism and consequence: as the band narrows as $1/\alpha$, trainability is lost.}
\label{fig:surrogate}
\end{figure}

Figure~\ref{fig:surrogate}A illustrates test accuracy as a function of the surrogate slope $\alpha=1/T_{\text{eff}}$. Accuracy peaks across a broad plateau at intermediate values of $\alpha$ before collapsing as $\alpha$ grows. This illustrates the relationship between trainability and backward temperature implied by Equation~\eqref{eq:surrogate_grad}, where $\alpha$ acts as the backward inverse temperature at which gradients are evaluated. At large $\alpha$, the backward temperature $T_{\text{eff}}=1/\alpha$ is low: the surrogate derivative concentrates into a narrow window around the firing threshold. Because the ordered-phase pinning pushes almost every attention pair away from this threshold, the majority of units receive vanishing gradients, causing learning to stall. Conversely, at the smallest values of $\alpha$, the backward temperature is high: the surrogate is broad, but it serves as a poor approximation to the hard forward map, introducing an architectural bias that slightly reduces accuracy. Trainability is therefore optimised at an intermediate backward temperature, revealing the surrogate slope as a genuine thermodynamic control variable rather than an arbitrary smoothing constant.

Figure~\ref{fig:surrogate}B elucidates the dynamic mechanism driving this collapse. It plots the active-band fraction (the proportion of attention pairs satisfying $|\gamma s_{ij}-\vartheta|<1/\alpha$ in the trained network) against $\alpha$ on logarithmic axes. If the pre-activation scores were statically and densely distributed around the threshold, this fraction would scale purely geometrically as $O(1/\alpha)$ with an exponent of $-1.00$. Instead, a power-law fit on the trained model reveals a steeper exponent of $-1.51$. This deviation provides direct empirical evidence for the ordered-phase pinning described by Equation~\eqref{eq:pinning}. During learning, the network actively shifts scores away from the trainable window, pushing them deep into the frozen regime to minimise entropy.

Together, these two panels separate the underlying thermodynamic mechanism from its macroscopic consequence. Panel B measures how the trainable band shrinks, reflecting a geometric narrowing of $1/\alpha$ compounded by the network's learned evacuation of that boundary. Panel A measures the resulting loss of trainability that this shrinking ultimately produces. The same ordered-phase pinning that renders the forward pass sparse and energy-efficient is exactly what dynamically places the operating point outside this vanishing-gradient band. Consequently, energetic efficiency and optimisation difficulty emerge as two faces of a single zero-temperature limit.

\subsection{Entropy dynamics and the cost of learning.}
\begin{figure}[t!]
\centering
\includegraphics[width=\textwidth]{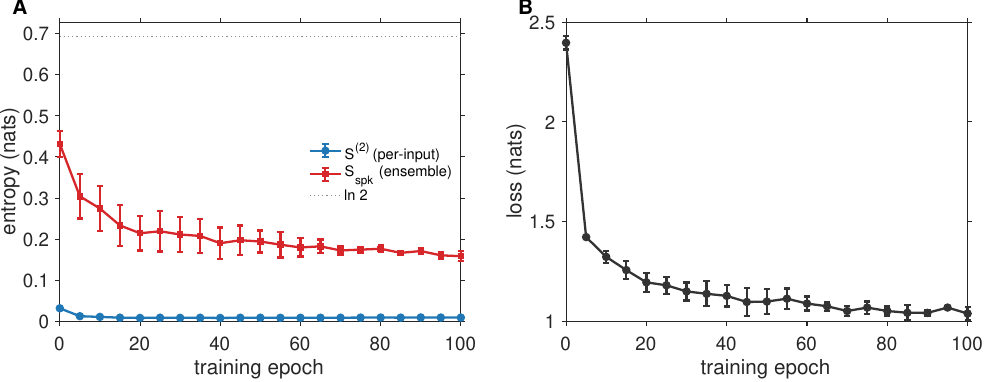}
\caption{Entropy dynamics of the spiking model during training on CIFAR-10, shown as the mean over $K$ seeds with error bars. \textbf{A}: the per-input two-level entropy $\bar S^{(2)}$ (blue), evaluated at the backward temperature $\beta_s=\alpha$, and the across-ensemble spike-pattern entropy $S_{spk}$ of Equation~\eqref{eq:sspk} (red), against training epoch; the dotted line is the maximum binary entropy $\ln 2$. The per-input entropy stays near zero for the whole run, so the attention pattern is frozen in the ordered phase at every epoch and the pinning of Equation~\eqref{eq:pinning} holds along the trajectory, whereas the across-ensemble entropy starts substantial and falls steadily, representing the informative reorganisation that learning performs. The across-seed spread of $\bar S^{(2)}$ is below the marker, as for the averaged observables of Figure~\ref{fig:relaxation}. \textbf{B}: the evaluation loss over the same window. The loss falls together with $S_{spk}$, placing the cost of learning in the across-ensemble spike pattern rather than in the per-input distribution.}
\label{fig:entropy}
\end{figure}

Figure~\ref{fig:entropy} tracks the two attention entropies along the training trajectory. Evaluated at the backward temperature $\beta_s=\alpha$, the local per-input two-level entropy $\bar S^{(2)}$ begins at a few hundredths of a nat and remains strictly flat for the entire run. The per-input attention pattern is therefore frozen in the ordered phase at every epoch, confirming that the pinning established in Equation~\eqref{eq:pinning} holds throughout the entire optimisation path. Conversely, the global across-ensemble spike-pattern entropy $S_{spk}$ exhibits the opposite behaviour: it begins near $0.43$ nats and falls steadily. Figure~\ref{fig:entropy}B demonstrates the evaluation loss falling in exact alignment with this early-dominated profile. This correspondence confirms that learning selectively tunes which attention neurons fire across different inputs, driving down $S_{spk}$, while the local per-input entropy contributes nothing because it is already pinned near zero. Across random seeds, $S_{spk}$ displays visible variation; in contrast, the spread of $\bar S^{(2)}$ is smaller than the plot markers because it is averaged over the entire attention array and remains securely pinned near zero.

These results provide the empirical validation for Equation~\eqref{eq:first_law}. The traditional softmax entropic term collapses because the local per-input entropy is permanently pinned near zero. Instead, the informative change shifts entirely to the global spike pattern, where the reduction $|\Delta S_{spk}|$ represents the quantity that the spike-generation First Law charges against the free-energy budget $-\Delta\mathcal{L}$. The figure confirms this qualitative framework by highlighting the frozen per-input entropy alongside the actively evolving spike-pattern entropy that our derivation theoretically separated. However, it does not explicitly test the quantitative Landauer bound derived in Equation~\eqref{eq:landauer_bound}. This theoretical floor is loose at the deployed backward temperature, and its strict empirical verification requires a precise calculation of the mutual-information work term, which we leave to future investigation.

\section{Discussion}
\label{sec:discussion}

In this paper, we developed a statistical-mechanical framework for softmax-free spiking attention, demonstrating that it operates exactly as the zero-temperature limit of a product of independent two-level systems. Removing the softmax layer shifts the underlying ensemble from a single coupled $N$-state categorical system to $N$ decoupled binary decisions. Under this framework, the training surrogate acts as the score gradient evaluated at a finite backward temperature defined by the surrogate slope $\alpha$ (Equation~\eqref{eq:correspondence}). This correspondence provides a mathematical basis for the defining properties of spiking attention. Specifically, the forward pass is strictly pinned to the ordered phase ($\phi^{(2)}\to1$) for every input (Equation~\eqref{eq:pinning}). This structural rigidity explains both the sparse, multiplication-free efficiency that motivates spiking architectures and the optimisation difficulty caused by an operating point trapped in a vanishing-gradient regime scaling as $O(1/\alpha)$. Because the standard entropic Landauer term collapses, the fundamental cost of learning shifts from sharpening a probability distribution to generating physical spikes, establishing an activity floor set by the per-spike energetic cost $C_{AP}$ (Equation~\eqref{eq:landauer_bound}).

Numerical experiments on CIFAR-10 confirm these theoretical predictions, placing spiking and softmax attention on a unified thermodynamic continuum. The trained spiking Transformer sits at the fully ordered endpoint of the two-level relaxation, exhibiting the interior Schottky anomaly characteristic of a pure two-level system. Tracked across the training trajectory, the local per-input entropy remains frozen at zero while the global across-ensemble spike-pattern entropy decays in alignment with the loss. This dynamic confirms the qualitative framework of the proposed First Law (Equation~\eqref{eq:first_law}). Furthermore, treating the surrogate slope as a backward temperature accurately predicts network trainability: accuracy peaks at intermediate temperatures and degrades severely as the active band geometrically narrows. These results establish a formal thermodynamic foundation for softmax-free attention, grounding its efficiency in the same binary systems that govern classical energy-based models. This perspective also recasts the surrogate slope as a physical control parameter, implying that training schedules which gradually cool toward the hard threshold act as the exact mathematical equivalent of simulated annealing.

While these findings validate the two-level correspondence (Equation~\eqref{eq:correspondence}), the ordered-phase pinning (Equation~\eqref{eq:pinning}), and the gradient support bounds, the physical postulate regarding the Landauer floor (Equation~\eqref{eq:landauer_bound}) requires further quantitative testing. This energetic floor assumes that the loss reduction acts as the free-energy budget for reorganising the global spike pattern. Because this theoretical bound is loose at the deployed backward temperature, a rigorous test requires evaluating the full First-Law balance (Equation~\eqref{eq:first_law}) by measuring the mutual-information work term $\Delta I(\hat Y;Y)$ directly along the training trajectory. Future work will focus on closing this quantitative balance and exploring hybrid operators that reintroduce partial cross-key normalisation, which could connect the zero-temperature spiking limit back to finite-temperature dynamics. Extending these measurements to larger architectures, language modelling, and native event-driven datasets will determine whether this backward-temperature framework consistently dictates trainability across domains. Ultimately, treating the surrogate slope rigorously as a physical temperature provides a theoretical lens for analysing any neural architecture that trains hard-threshold units via continuous surrogate gradients.

\section*{Data and Code Availability}
The code and models used to produce the results in this paper are available from the corresponding author upon reasonable request.

\section*{Acknowledgements}
This work was partially supported by grant RGPIN-2022-03042 from Natural Sciences and Engineering Research Council of Canada.

\printcredits

\bibliographystyle{elsarticle-num}
\bibliography{cas-refs}

\end{document}